\documentclass[lettersize,journal]{IEEEtran}
\usepackage{amsmath,amsfonts}
\usepackage{algorithmic}
\usepackage{array}
\usepackage[caption=false,font=normalsize,labelfont=sf,textfont=sf]{subfig}
\usepackage{textcomp}
\usepackage{stfloats}
\usepackage{url}
\usepackage{verbatim}
\usepackage{graphicx}
\usepackage{booktabs}
\usepackage{amsmath} 
\usepackage{setspace}
\usepackage{amssymb}
\usepackage{pifont}
\usepackage{xspace}
\usepackage{xcolor}
\usepackage{placeins}

\newcommand{\cmark}{\ding{51}}
\newcommand{\xmark}{\ding{55}}

\def\BibTeX{{\rm B\kern-.05em{\sc i\kern-.025em b}\kern-.08em
    T\kern-.1667em\lower.7ex\hbox{E}\kern-.125emX}}
\usepackage{balance}
\begin{document}
\title{Top-Down Viewing for Weakly Supervised Grounded Image Captioning}
\author{Chen Cai, Suchen Wang, Kim-hui Yap,  Yi Wang
\thanks{Corresponding author: Kim-Hui Yap.}
\thanks{Chen Cai, Suchen Wang and Kim-hui Yap are with the School of Electrical and Electronic Engineering, Nanyang Technology University, Singapore (email:E190210@e.ntu.edu.sg; ekhyap@ntu.edu.sg). Yi Wang with the Department of Electrical and Electronic Engineering, The Hong Kong Polytechnic University.}
}

\markboth{Preprint}%
{How to Use the IEEEtran \LaTeX \ Templates}

\maketitle

\begin{abstract}
Weakly supervised grounded image captioning (WSGIC) aims to generate the caption and ground (localize) predicted object words in the input image without using bounding box supervision. Recent two-stage solutions mostly apply a bottom-up pipeline: (1) first apply an off-the-shelf object detector to encode the input image into multiple region features; (2) and then leverage a soft-attention mechanism for captioning and grounding. However, object detectors are mainly designed to extract object semantics (i.e., the object category). Besides, they break down the structural images into pieces of individual proposals. As a result, the subsequent grounded captioner is often overfitted to find the correct object words, while overlooking the relation between objects (e.g., what is the person doing?), and selecting incompatible proposal regions for grounding. To address these difficulties, we propose a one-stage weakly supervised grounded captioner that directly takes the RGB image as input to perform captioning and grounding at the top-down image level. In addition, we explicitly inject a relation module into our one-stage framework to encourage the relation understanding through multi-label classification. The relation semantics aid the prediction of relation words in the caption. We observe that the relation words not only assist the grounded captioner in generating a more accurate caption but also improve the grounding performance. We validate the effectiveness of our proposed method on two challenging datasets (Flick30k Entities captioning and MSCOCO captioning). The experimental results demonstrate that our method achieves state-of-the-art grounding performance. The code for this work will be made publicly available.
\end{abstract}


\section{Introduction}
Image captioning is a fundamental problem in computer vision that recognizes the objects and the relationships in the image and describes them with natural language \cite{M2, clipcap, PrefixClip,RFLSTM, ShowAttenTell, TAcap}. More recently, the bottom-up attention-based mechanism \cite{BUTD} has been widely adopted for image captioning framework and achieves remarkable success \cite{APcap, Multicap, AOA, XTran, ObjectToWords}.

\begin{figure}[t!]
\begin{minipage}[b]{1.0\linewidth}
    \centering
    \centerline{\includegraphics[width=8.5cm]{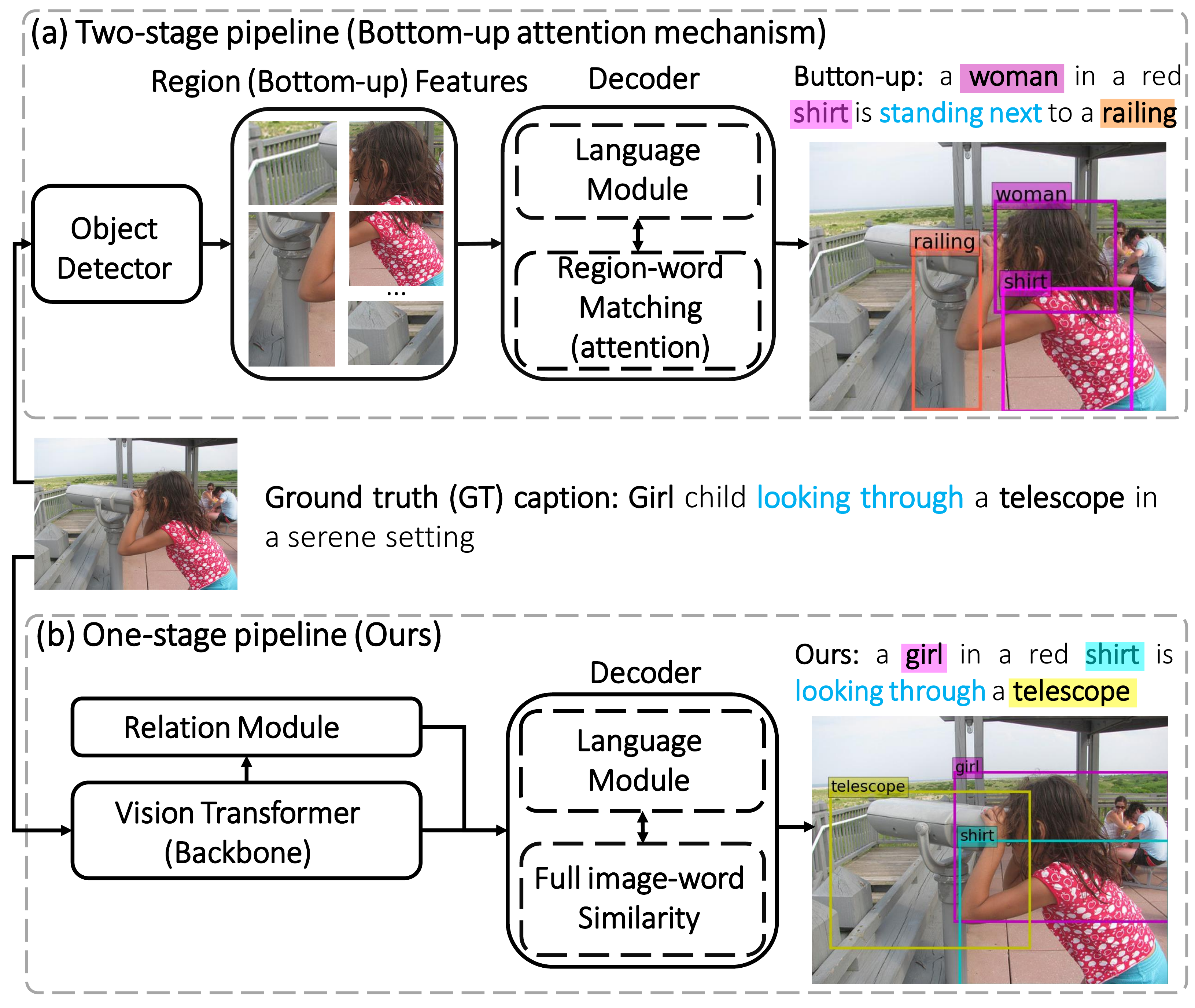}}
    \caption{(a) \textbf{Two-stage pipeline}: use object-focused region (bottom-up) features and soft-attention \cite{BUTD} for captioning and grounding. (b) \textbf{One-stage framework (ours)}: use raw RGB image as input for captioning and grounding. Instead of selecting salient region features, the one-stage method allows us to calculate similarity metrics between words and the entire image for grounding. Besides, we explicitly model the relation semantics (e.g., ``looking though") and utilize them as contextual information to assist in predicting the desired groundable (e.g., ``telescope") words in the caption.
    }
    \label{fig:fig1}
\end{minipage}
\end{figure}

Apart from the significant advances achieved in image caption generation, many mainstream works \cite{MultiGIC,VC,GCNGIC,GVD,SCAN} explore more grounded image captioners that localize the groundable object words while generating a caption to facilitate the interpretability of image captioning models.  These methods develop regularization schemes to match the object words in the generated caption with region features and use the corresponding coordinates as grounding regions. Some previous methods \cite{GCNGIC, GVD} utilize bounding boxes annotation of each groundable object word at the training stage and achieve satisfactory grounding performance. However, the cost of annotating bounding boxes for large-scale captioning datasets is extortionate. Besides, this group of methods is inherently unsuitable for handling relation words since the verb words are ambiguous to be annotated by bounding boxes. Recently, weakly supervised grounded image captioning (WSGIC) methods \cite{Prophet, Cyclical, SCAN} have drawn more attention. It alleviates the requirement of box annotations and learns to ground only based on the given image and captioning pairs during training. Nevertheless, most existing methods use two-stage pipelines that raises three difficulties: 1) limiting the model's adaptability and efficiency in a real-world application. The raw RGB images must process with region proposal operation (i.e., RPN) in object detection to extract region features in the first stage. 2) The input region features for caption generation care more about object class information which overlooks the benefits of relation semantic information for caption generation and grounding. 3) Losing the global understanding of images leads to aligning incompatible regions for grounding. The grounded captioners tend to select the most discriminative region feature as the grounding region which inferior grounding performance. For instance, the two-stage method tends to select the ``head'' proposal region as the localized region for the groundable word ``woman'' in the caption rather than ground the entire body in Figure \ref{fig:fig1}.

Furthermore, the existing WSGIC methods focus more on object word generation using object-focused region features. The benefits of emphasizing indispensable relation semantic information in the grounded caption are largely underexplored. As shown in Figure \ref{fig:fig1}, we observe that existing bottom-up methods are more sensitive to recognizing the objects in the image (e.g., woman, shirt, railing, etc.) while often overlooking the actions, behaviors, and activities (e.g., looking through, etc.), which can results in caption hallucinations problem \cite{SCAN, Cyclical} and lead to inferior captions. For example, with less informative relation words ``standing next'' as context information, the captioner tends to generate ``railing'' for grounding rather than the more desired word ``telescope'' in the ground truth. We observe that relation words often serve as a context that benefits object word generation in caption modeling.

To address these problems, we propose a one-stage grounded captioner that can be trained in a weakly supervised manner. Different from previous approaches using pre-computed region features, we eliminate the off-the-shelf object detector and directly take the raw RGB image as input. It allows the captioner to perform captioning and grounding on the basis of the entire image. Specifically, our top-down vision transformer-based \cite{VIT} encoder primarily encodes the raw image to produce the class token and visual patch token representations. The grounded language decoder then utilizes these visual representations to compute word representations at each time step. Concurrently, the proposed Recurrent Grounding Module (RGM) in the decoder takes the word representations and the visual features to compute the Visual Language Attention Maps (VLAMs) for grounding. The VLAMs present the spatial region and location of the generated groundable object words in the caption. Besides, as we model at the image level representation rather than isolated region features, it allows us to explore how to incorporate relation semantic information in one-stage grounded image captioning. In this work, we introduce a [REL] token to capture relation semantic information that helps the grounded captioner generate the more desired relation and object words in the caption. Our study shows that incorporating relation semantic features into the formulate can increase the captioning and grounding quality.    

The main contributions of our paper can be summarized as follows: 
\begin{itemize} 
\item We propose a one-stage weakly supervised grounded image captioning method to generate image captions and perform grounding in a top-down manner.

\item We introduce a relation token that models relation semantic information, which provides rich context information to grounded captioners to significantly benefit the generation of captions.

\item We proposed a recurrent grounding module that enables the grounded captioner to compute precise Visual Language Attention Maps (VLAMs) for the object words, enhancing the grounding performance.

\item We achieve state-of-the-art grounding and competitive captioning performance in the Flick30k Entities and MSCOCO captioning datasets.
\end{itemize}

The remaining sections are organized as follows. Section II describes related work on grounded image captioning, visual grounding, and weakly supervised object localization. In Section III, we introduce the proposed top-down image encoder and the grounded language decoder. In Section IV, we show the extensive experimental results of the Flick30k Entities captioning and MSCOCO captioning datasets. Finally, we conclude this work in Section V.

\section{Related Works}
\subsection{Grounded image captioning}
Many grounded image captioning models utilize pre-trained object detector to extract region features and adopt attention mechanisms to accomplish distinct advances in grounding and captioning. Zhou et al. \cite{GVD} utilize the aligned bounding box annotations of the noun phases to improve the captioning and grounding quality.
Zhang et al. \cite{GCNGIC} further extract two-stage scene graph relation features for the supervised grounded captioning.
However, labeling and aligning the bounding boxes with noun words in the captions is expensive for the large-scale dataset.  
Recent methods \cite{MultiGIC, Prophet, Cyclical, SCAN} have shown a promising alternative to improve the grounding accuracy in a weakly supervised manner. Ma et al. \cite{Cyclical} proposed the cyclical training regimen that first localizes the region of interest (ROI) of the generated words using a language decoder, then reconstructs the sentence based on localized visual regions to regularize the captioning model. Liu et al. \cite{Prophet} explored the prophet attention that utilizes future information to compute ideal attention weights to regularize the deviated attention. Chen et al. \cite{MultiGIC} studied an ensemble training strategy to solve incompatible region-word alignment problems. It fuses ROIs produced by multiple grounded captioners as the final grounded region. In addition, recent works \cite{SCAN, VC, CVAE} have developed some pre-trained/pre-built weak supervision techniques to encourage the grounded captioner to achieve a better word-region alignment. Different from pioneer works that are studied upon the pre-computed region features, we aim to explore a one-stage WSGIC solution. We expect that the captioner can learn how to generate captions and locate the groundable object words directly from raw images in a weakly supervised manner. Furthermore, we explore a simple and effective method to incorporate semantic relations into a one-stage grounded captioner.

\begin{figure*}[t!]
    \begin{center}
        \centering
        \includegraphics[width=18cm]{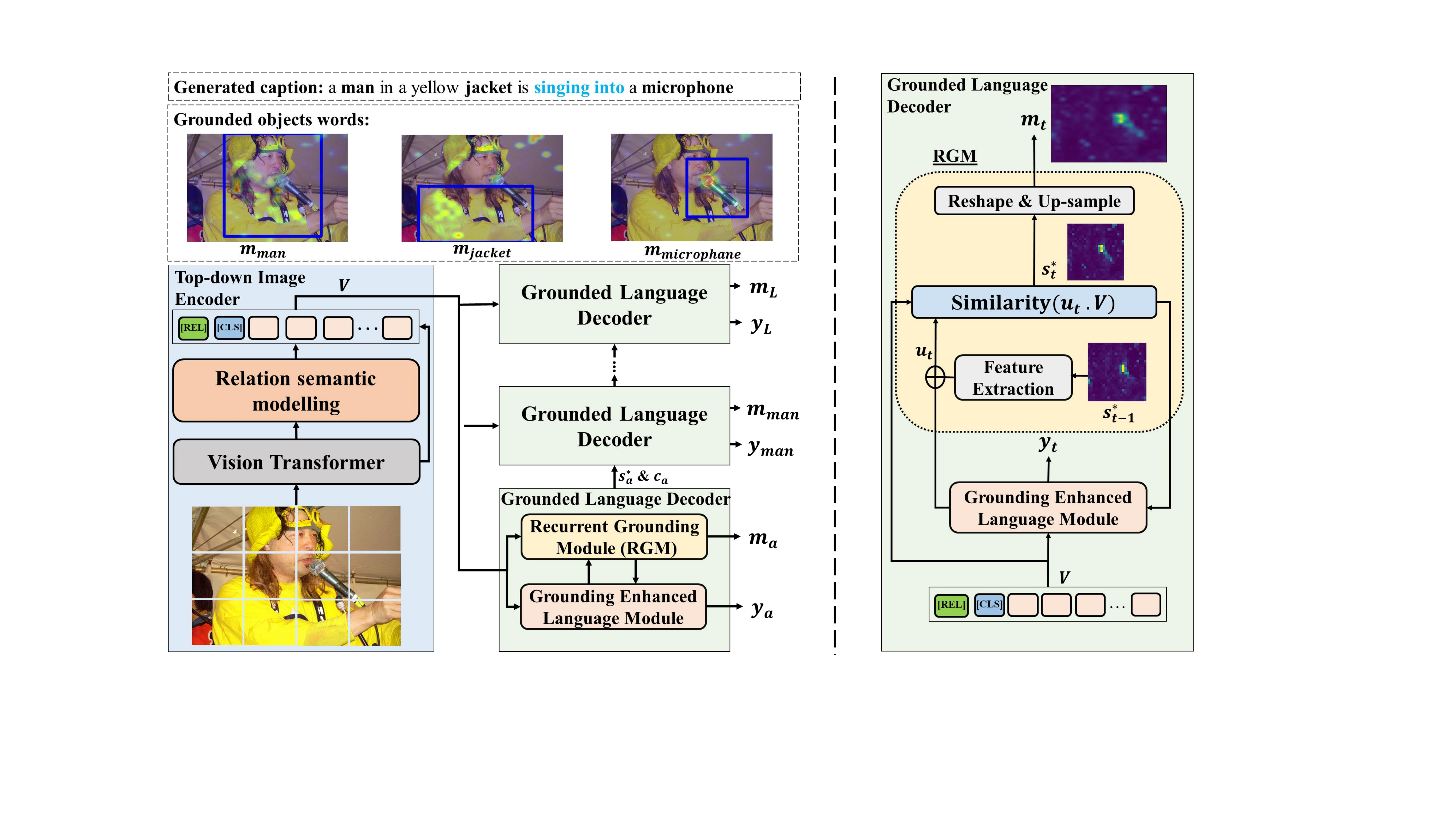}
        \caption{Overview of the proposed weakly supervised grounded image captioning model. It consists of a top-down image encoder and a grounded language decoder. The image encoder uses the Vision Transformer backbone to encode the RGB raw image into [CLS] and patches token representations. A new [REL] token is concatenated with frozen patch representations and trained to model the relation semantic information. The decoder utilizes these visual representations as input to generate the caption $\mathbf{Y}$, and computes the Visual-Language Attention Maps (VLAMs) $\mathbf{M}$ for localization based on dot product similarity between visual $\mathbf{V}$ and word $\mathbf{u}_t$ representations. The top-left part of the figure shows the generated caption, and the grounded object regions computed using the VLAMs (e.g., $\mathbf{m}_{jacket}$). }
        \label{fig:fig2}
        \vspace{-6pt}
    \end{center}
\end{figure*}

\subsection{Visual grounding}
VG task aims to localize the object region in an image based on a natural language phrase or sentence query. The pioneer works \cite{VGAA, RVG-Tree, RCCF} researches on the fully supervised VG, which aligns the noun phrases and extracted regional features via decoder using annotated bounding box supervision. More recently, \cite{TransVG, VLVR} proposed to ground language query to the corresponding region in the image in an end-to-end manner to increase the adaptability of the VG model. Besides supervised works, some works \cite{MLMCSS, KGPR, ContrasVG} explored weakly supervised VG that learns from image-sentence pairs without the need of bounding box supervision. Akbari et al. \cite{MLMCSS} explored the non-linear mapping of visual and textual features at multiple levels, and it guides the model using a multi-level multimodal attention mechanism for visual grounding.  Wang et al. \cite{ContrasVG} improved visual grounding performance via contractive learning \cite{ContracCode}. Liu et al. \cite{KGPR} proposed a knowledge-guided pairwise reconstruction network that models the relationships between image-sentence pairs and performs localization. 
Unlike the VG task, our task only takes images as input and aims to automatically determine the most important visual concepts to describe.

\subsection{Weakly supervised object localization}
WSOL focuses on localizing objects with image-level category labels. Most WSOL methods \cite{AttDropOut, TS_CAM, Vitol, BGL, CAM, CPlocal} use the class activation map (CAM) to indicate object region with respect to the prediction of class label. Choe et al. \cite{AttDropOut} propose an attention-based dropout layer that exploits the attention mechanism to process the feature maps and improve the localization quality. The work \cite{erasLoc} introduces erasing integrated learning that Investigates both the less discriminatory area and the high response class-specific area to explore the complete extent of the object region. Choe et al. \cite{TS_CAM} model the long-range dependency over image patches through Vision Transformer \cite{VIT}, and they proposed a token semantic couple attention map (TS-CAM) to solve partial activation issues. To enhance localization accuracy, Gupta et al. \cite{Vitol} introduced a patch-based attention dropout layer. In contrast to WSOL, our work performs localization by exploring the attention map generated via the dot product similarity between visual representations and generated caption words.

\section{Methodology}
Figure \ref{fig:fig2} gives an overview of the proposed WSGIC model. In this work, we adopt an encoder-decoder framework. It is composed of a \textit{top-down image encoder} to encode the input RGB image with object and relation semantic information, and a \textit{grounded language decoder} to recurrently generate the caption and ground the objects in the image. For each time step, our grounded language decoder outputs the caption word {$\mathbf{y}$} (e.g., $\mathbf{y} = man, jacket$, etc.) and computes a Visual Language Attention Map (VLAM) $\mathbf{m} \in\mathbb{R}^{H \times W}$ (e.g., $\mathbf{m}_{man}, \mathbf{m}_{jacket}$, etc.) to localize the groundable word with bounding box coordinate $\mathbf{b}_t=\{x_1, y_1, x_2, y_2\}$. In the following subsection, we elaborate the details of our top-down image encoder and grounded language decoder.

\subsection{Top-down Image Encoder}
The image encoder encodes a fixed resolution (e.g., $224 \times 224$, $384 \times 384$) RGB image $\mathbf{I} \in\mathbb{R}^{H\times W\times 3}$ with Vision Transformer (e.g., Vit \cite{VIT}, DeiT \cite{DEIT}) pre-trained model. Concretely, the image is first divided into $P \times P$ patches and encoded as a sequence of patch embeddings $\mathbf{Z}_{\text{patch}}^0 = [\mathbf{z}_1^0; \mathbf{z}_2^0; \dots; \mathbf{z}_N^0] \in \mathbb{R}^{N \times D}$, where $\mathbf{z}_i^l \in \mathbb{R}^D$ represents the $i$-th patch token that has dimension $D$, $N = \frac{H}{P} \times \frac{W}{P}$ and $l \in \{0, 1, ..., L\}$ denotes the total number of patch tokens and Transformer layers, respectively. Usually, a learnable [CLS] token $\mathbf{z}_{\text{cls}}^0 \in \mathbb{R}^D$ is prepended and trained along with image patches using $L$ transformer layers to capture useful object information. Therefore, we treat the output $\mathbf{z}_{\text{cls}}^L$ as an object semantic representation and $\mathbf{Z}_{\text{patch}}^L$ as patch representations. Taking advantage of the [CLS] token representation can drive the caption decoder to emphasize and generate desired object words.

\subsubsection{Visual relation semantic modeling} Besides, letting the captioning model describe the relationship between objects is no less important. Similar to the [CLS] token, we introduce a new learnable [REL] token and expect it to capture image-level relation semantic information. Specifically, let $\mathbf{z}_{\text{rel}}^0 \in \mathbb{R}^D$ be the [REL] token, then we concatenate it with the frozen patch representations $\mathbf{Z}_{\text{patch}}^L$ and train it with additional $L_r$ number of learnable transformer encoder layers $\text{T}_{enc}$, which can be written as:
\begin{equation}\label{eq:eq1}
\begin{split}
    \mathbf{z}_{rel}^{Lr}, \mathbf{Z}_{\text{patch}}^{(L+L_r)} = \text{T}_{enc}([\mathbf{z}_{rel}^{0}, \mathbf{Z}_{\text{patch}}^{L}])
\end{split}
\end{equation}
Here [,] denotes concatenation. To ease the presentation, we omit the multi-head attention (MHA) \cite{Trans}, position-wise Feed-Forward Networks (FFN), and Layer Normalization operation (LN) \cite{Trans} in the equation \eqref{eq:eq1}. $\mathbf{z}_{rel}^{Lr}$ will then inject into prediction heads (e.g., MLP) and train to capture the semantic information of selected relation classes similar to multi-label image classification. 
These representations are fused and projected into $d=512$ dimension with a linear layer, whereas we use $\mathbf{V} = [\mathbf{z}_{\text{rel}}^{Lr}; \mathbf{z}_{\text{cls}}^L; \mathbf{Z}_{\text{patch}}^L ] \in \mathbb{R}^{(N+2) \times d}$ to represent the final output of our top-down image encoder. The output $\mathbf{V}$ will be passed into the subsequent caption decoder. Our intention is to inject the class and relation semantic representation into the caption decoder to benefit captioning and grounding performance.

\subsubsection{Selection of relation classes} To model the relation semantics with relation labels. We utilize both verb (e.g., playing, jumping, etc.) and preposition (e.g., across, above, etc.) words in the caption as relation labels similar to VG dataset \cite{VG} with the help of Part-of-Speech (POS) tagging. By combining both types of words, relation modeling can capture a wider range of relationships between objects in the image, making it more accurate and informative. Additionally, not all captions contain verbs, so including prepositions can help ensure that all images are associated with a relation label for classification. For instance, 1) given a ground truth caption of \textit{A person in front of a building}, the proposed method tends to model the relation semantic of \textit{in front of} (prepositions) to assist the caption generation. 2) proposed method can model relation semantic concepts such as \textit{playing with} (verb and preposition) when the caption is \textit{The girl is playing with her dog}. The most frequently used relation words (62 and 72 relation classes for Flickr30k-Entities and MSCOCO, respectively) are selected to ensure every image is associated with at least one relation label for relation modeling (see Figure \ref{fig:flickS} and \ref{fig:cocoS} for the statistics of relation labels). 

\begin{figure*}[t!]
    \begin{center}
        \centering
        \includegraphics[width=17cm]{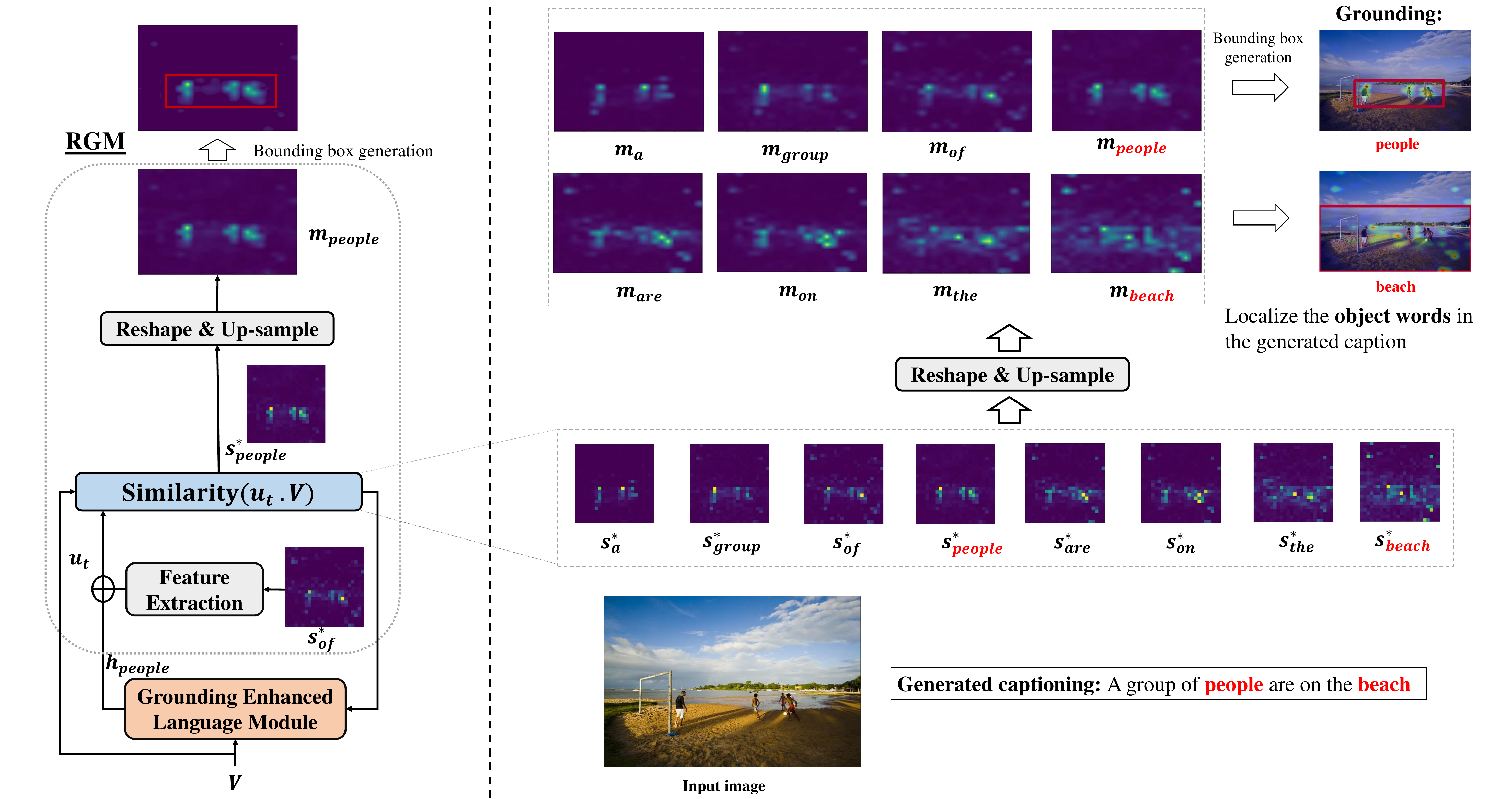}
        \caption{The illustration of the Recurrent Grounding Module (RGM) and grounding process. We enhance the similarity attention metrics $\mathbf{s}_{people}^*$ ($\mathbf{s}_{t}^*$) for the current time step recurrently by conditioning on the $\mathbf{s}_{of}^*$ ($\mathbf{s}_{t-1}^*$) from the previous time step. This enables us to compute more precise $\mathbf{m}_{people}$ for grounding (shown in Figure \ref{fig:fig9}). On the right of the figure, we show the $\mathbf{s}^*$ that is computed concurrently with each generated word. During the testing stage, we reshape \& up-sample the  $\mathbf{s}^*$ to $\mathbf{m}$, and localize the groundable objects words using $\mathbf{m}_{people}^*$ and $\mathbf{m}_{beach}^*$.
        }
        \label{fig:rgm}
    \end{center}
\end{figure*} 

\subsection{Grounded Language Decoder}
The proposed grounded caption decoder consists of a grounding enhanced language module and a recurrent grounding module (RGM). The language module generates caption representation $\mathbf{C} = [\mathbf{c}_1; \mathbf{c}_2; \mathbf{c}_3; \dots; \mathbf{c}_L]$ sequentially, which can be used to predict the caption words $\textbf{Y}$, where $\mathbf{c}_i \in \mathbb{R} ^ d$, $L$ is the length of generated caption and $d=512$ is the dimension of caption representation. The RGM computes the Visual-Language Attention Maps  (VLAMs) $\mathbf{M} = [\mathbf{m}_1; \mathbf{m}_2; \mathbf{m}_3; \dots; \mathbf{m}_L]$ for grounding sequentially with the generation of caption words, where  $\mathbf{m}_i \in \mathbb{R}^{H \times W}$.

\subsubsection{Recurrent grounding module}
For the time step $t$, assume we are generating the $\mathbf{m}_{people} \in \mathbb{R}^{H \times W} $ (e.g., in Figure \ref{fig:rgm}) that represents the spatial region of the generated object word ``people". We first need to generate the word and visual similarity attention metrics $\mathbf{s}_{people}^* \in \mathbb{R} ^ {N}$, which can be computed through dot product between word representation $\mathbf{h}_{people} \in \mathbb{R} ^ {d}$ and visual representation $\mathbf{V}$, where $\mathbf{h}_{jacket}$ is the output of the LSTM sub-module from the \textbf{grounding enhanced language module}, and $N = \frac{H}{P} \times \frac{W}{P}$ (e.g., $N=24 \times 24$).

Moreover, similar to sequential modeling, we propose to enhance the similarity attention metrics $\mathbf{s}_{t}^*$ (e.g., $\mathbf{s}_{people}^*$) for the current time step by conditioning on the $\mathbf{s}_{t-1}^*$ (e.g., $\mathbf{s}_{of}^*$) from the previous time step. This enables us to compute more precise $\mathbf{M}_{people}$ for grounding (shown in Table \ref{tab:tab2} and Figure \ref{fig:fig9}). We view $\mathbf{s}_{t-1}^*$ as the simple visual representation containing the information of a pseudo ``object'' with the pixel size of $N$. We then extract its features and project it into dimension $d$ using a single dense layer. In particular, we perform the following steps to compute $\mathbf{m}_{t}$ for the current time step:
\begin{flalign}
   & \mathbf{m}_{t} = \Phi\{\mathbf{s}_{t}^*\} \in \mathbb{R}^{H \times W} \\ \label{eq3}
   & \mathbf{s}_{t}^* \in \mathbb{R} ^ N = \delta\{\mathbf{s}_{t}\} \in \mathbb{R} ^ {N+2} =  \text{Softmax}(\frac{\mathbf{u}_t \mathbf{V}^\top}{\sqrt{d}})\\ 
   & \mathbf{u}_t = [\mathbf{h}_t + f(\mathbf{s}_{t-1}^*)] \in \mathbb{R} ^ d
\end{flalign}
Where $f(.)$ denotes dense layers, and $\top$ represents the transpose operation. 
$\mathbf{u}_{t}$ (Query) is the fusion representation that fuses the word and the representation of $\mathbf{s}^*_{t-1}$. $\mathbf{s}_t \in \mathbb{R}^{(N+2)}$ has the same patch size as $\mathbf{V}$ (Key) after dot product and Softmax function. Here, the split operation $\delta\{.\}$ is introduced to split $\mathbf{s}_t \in \mathbb{R}^{(N+2)}$ to $\mathbf{s}_t^* \in \mathbb{R}^{N}$ ($N = \frac{H}{P} \times \frac{W}{P}$) implying how much attention the word representation paid for each visual patch token, where $+2$ is the attention score of [CLS] and [REL] token. $\Phi\{.\}$ denotes the reshaping and up-sampling process to convert and enlarge the similarity attention matrices $\mathbf{s}_t^*$ to VLAM of $\mathbf{m}_t\in \mathbb{R}^{W \times H}$. The generation of $\mathbf{s}_t^*$ can be extended to various numbers of the parallel heads $N_{h}$ and perform summation over $\{{\mathbf{s}_t^*}\}_{i=1}^{N_{h}}$ to compute more accurate VLAM and improve the grounding performance (illustrated in Table \ref{tab:tab4}).

\subsubsection{Object bounding box generation}
The VLAMs are able to highlight the discriminative region of the object words (demonstrated in Figure \ref{fig:rgm}). We borrow the idea of the thresholding approach \cite{thresholding} to find the object-bounding box based on generated VLAMs during the testing stage. Given the $\mathbf{m}$ (e.g., $\mathbf{m}_{people}$) that has the same image size $H \times W$ as input image. We define a binarized mask $K \in \{0, 1\}^{W\times H}$, where $K_{x, y}=0$ if the pixel at position $x,y$ belongs to background region,  $K_{x, y}=1$ if it is part of the object (foreground) regions. The produced masks can be computed as:
\begin{equation}
\begin{split}
  \mathbf{K}_{x, y} =\left\{
  \begin{array}{@{}ll@{}}
    1, & \text{if} \ \mathbf{m}^{x, y} > \rho, \ 0 <  \rho < 1 \\
    0, & otherwise. \\
  \end{array}\right.
\end{split}
\end{equation} 

Where the threshold $\rho=0.05$ is used to differentiate the region of the object and background in the $\mathbf{m}_t$. With the shape and location information presented in $\mathbf{K}$, we are able to determine the bounding box coordinates $\mathbf{b}_t=\{x_1, y_1, x_2, y_2\}$ of the object.

\subsubsection{Grounding enhanced language module} Generating accurate word representations is crucial in computing correct VLAMs for grounding, especially for the groundable object words in the caption. These word representations are used to produce precise word-visual similarity attention matrices (eq. \eqref{eq3}), which are essential for achieving the correct localization of objects. Hence, in this work, we utilize the relation and object semantic representations as contextual cues to assist the language model in generating desired word representations, which leads to benefits in both captioning and grounding (shown in Table \ref{tab:tab2}). 

Motivated by many existing captioning works \cite{AOA, VITCAP, ObjectToWords}, the word representations are commonly generated with the Transformer-based decoder \cite{Trans} or RNN-based decoder \cite{LSTM}. Based on experiments, we discovered that utilizing the LSTM network is better suited for generating the word representations that are used to compute VLAMs for our one-stage WSGIC (shown in Table \ref{tab:tab6}). Therefore, the LSTM network is chosen to generate the word representations  $\mathbf{H} = [\mathbf{h}_1; \mathbf{h}_2; \mathbf{h}_3; \dots; \mathbf{h}_L] \in \mathbb{R}^{L \times d}$ based on visual features $\mathbf{V}$, that containing visual patch, [CLS] and [REL] semantic representations. Specifically, for each time step:
\begin{flalign}
   & \mathbf{c}_{t} = \text{LN}(\text{FFN}(\text{GLU}(\mathbf{h}_t + \mathbf{c}_t^g))) \in \mathbb{R}^d \\
   & \mathbf{c}_{t}^g = \mathbf{s}_{t}\mathbf{V} \in \mathbb{R}^d \\
   & \mathbf{h}_t = \text{LSTM}([\mathbf{\bar{v}} + \mathbf{c}_{t-1}, \mathbf{W}_e\Pi_t], \mathbf{h}_{t-1}) \in \mathbb{R}^d
\end{flalign}
Where $\mathbf{\bar{v}}$ denotes the mean pooled $\mathbf{V}$. $\mathbf{W}_e\in\mathbb{R}^{E\times|\sum|}$ is a word embedding matrix for a vocabulary $\sum$, $\Pi_t$ is the one-hot encoding of the input word at the current time step t and $[,]$ denotes concatenation. 
The $\mathbf{c}_{t}^g$ captures attended language representation that aggregates the visual tokens, object $\mathbf{z}_{\text{cls}}$ and relation  $\mathbf{z}_{\text{rel}}$ features through attention score $\mathbf{s}_{t}$ from equation \eqref{eq3} in GRM, which aids the caption generation and allows GRM to be optimized at the training stage. We use the gated mechanism like GLU \cite{GLU} to enhance the output representations for caption generation. The layer normalization LN and feedforward FFN sub-layer are added after the GLU sub-layer. The caption representations $\mathbf{C}$ are fed into a linear projection layer and Softmax layer for the prediction of words $\mathbf{Y}$.

\subsection{Training and Objectives}
In this work, we adopt the standard multi-label classification loss for relation prediction:
\begin{equation}
\begin{split}
   \mathcal{L}_{MLC} = -\sum_{i=1}^{N_c} z_i \log(p_i), p, z  \in \mathbb{R}^{N_c}, z \in \{1, 0\}
\end{split}
\end{equation}

Where $p$ and $z$ denote the prediction and ground-truth relation label respectively, $N_c$ is the number of relation class. We train the grounded image captioning model by optimizing the cross-entropy (XE) loss $L_{XE}$:
\begin{equation}
\begin{split}
  \mathcal{L}_{XE} = -\sum_{t=1}^{T} \log(p_\theta(\mathbf{y}_t^* | \mathbf{y}_{1:t-1}^* ))
\end{split}
\end{equation}
The model is trained to predict the target ground truth caption  $\mathbf{y}_{t}^*$ with the given words $\mathbf{y}_{1:t-1}^*$. The overall learning objective is $\mathcal{L} = \mathcal{L}_{MLC} + \mathcal{L}_{XE}$.

\section{Experiment}
\label{sec:sec4}
We evaluate our proposed weakly supervised grounded image captioning method on the popular Flickr30k-Entities \cite{Flickr30k} and MSCOCO captioning dataset. The flickr30k-Entities dataset consists of 31k images, and each image is annotated with 5 captions. It contains 275k annotated bounding boxes associated with natural language phrases and a total of 480 object classes. We selects 72 most frequently used relation classes in the dataset. Similar to the existing method \cite{Cyclical, Prophet}, we used splits from Karpathy et al. \cite{Karpathy}, where 29k/1k/1k images are used for training, validation, and testing, respectively. We drop the words that occur less than 5 times which end up with a vocabulary of 7000 words. For the MSCOCO caption dataset \cite{MSCOCO},  113k, 5k, and 5k images are used for training, testing, and validation, respectively. The vocabulary size for the COCO caption dataset is 9487. We select the 62 most frequently used relation classes in the COCO caption dataset for relation semantic modeling. Besides captioning quality, we evaluate the grounding quality on the MSCOCO dataset for 80 object categories. We merge the MSCOCO captioning Karpathy et al. \cite{Karpathy} testing split with the MSCOCO detection datasaet, which resulting 3648 image-caption-bounding boxes pairs to test the grounding performance.

We use the $\text{F1}_{all}$ and $\text{F1}_{loc}$ metrics that defined in GVD \cite{GVD} to evaluate the grounding quality. In $\text{F1}_{all}$, a region prediction is considered correct if the predicted object word is correct and the IoU between the predicted bounding box and ground truth is greater than 0.5. The $\text{F1}_{loc}$ mainly considers correctly predicted object words. More detailed information about $\text{F1}_{all}$ and $\text{F1}_{loc}$ metrics can be found in the supplementary material of GVD. We use standard evaluation metrics, including BLEU \cite{bleu}, METEOR \cite{meteor}, CIDEr \cite{cider} and SPICE \cite{Spice}, to evaluate the caption quality and compare with existing methods.

\begin{table*}[t!]
\setlength{\tabcolsep}{14pt}
\centering
  \caption{Performance comparisons on Flickr30k-Entities test split. The RL denotes that models are further trained using Reinforcement Learning. (Sup.) indicates the model is trained with bounding box supervision. All values are in percentages (\%), and higher is better. The symbols B, M, C, and S are the short-form of BLEU, METEOR, CIDEr, and SPICE scores, respectively. \textcolor{red}{Red} number indicates the best result of using XE loss, \textcolor{blue}{blue} number denotes the best in RL optimization.}
    \begin{tabular}{l|c|ccccc|cc}
    \hline
    Methods & RL &\multicolumn{5}{c|}{Caption eval}  & \multicolumn{2}{c}{Grounding eval} \\
    \hline
          & &\multicolumn{1}{c}{B@1} & \multicolumn{1}{c}{B@4} & \multicolumn{1}{c}{M} & \multicolumn{1}{c}{C} & \multicolumn{1}{c|}{S} & \multicolumn{1}{c}{\text{$\text{F1}_{all}$}} & \multicolumn{1}{c}{\text{$\text{F1}_{loc}$}} \\
    \hline
    GVD (Sup.)\cite{GVD}    &  \xmark & 69.9  & 27.3  & 22.5  & 62.3  & 16.5  & 7.55  & 22.2 \\
    \hline
    GVD \cite{GVD}    &  \xmark & 69.2  & 26.9  & 22.1  & 60.1  & 16.1  & 3.88  & 11.7 \\
    Cyclical \cite{Cyclical}  & \xmark  & 68.9  & 26.6  & 22.5  & 60.9  & 16.3  & 4.85  & 13.4 \\
    Prophet \cite{Prophet}  & \xmark & - & 27.2  & 22.3  & 60.8  & 16.3  & 5.45  & 15.3 \\
    $\text{CVAE}^{\dagger}$ \cite{CVAE} & \xmark  & - & 24.0  & 21.3  & 55.3    & 15.7  & 6.70  & 19.2 \\
    Ours  & \xmark & \color{red}\textbf{69.9}  & \color{red}\textbf{27.8}  & \color{red}\textbf{22.7} & \color{red}\textbf{64.1} & \color{red}\textbf{17.2} & \color{red}\textbf{7.88} & \color{red}\textbf{19.9} \\
    \hline
    $\text{SCAN}^{\dagger}$ \cite{SCAN} & \cmark  & 73.4 & 30.1  & 22.6  & 69.3    & 16.8  & 7.17  & 17.5 \\
    $\text{CVAE}_{RL}^{\dagger}$ \cite{CVAE} & \cmark  & - & 29.8  & 23.1  & 67.6    & 17.2  & 6.94  & 17.6 \\
    $\text{Ours}_{RL}$  & \cmark & \color{blue}\textbf{73.7}  & \color{blue}\textbf{30.8}  & \color{blue}\textbf{23.2} & \color{blue}\textbf{70.6} & \color{blue}\textbf{17.6} & \color{blue}\textbf{7.56} & \color{blue}\textbf{18.2} \\
    \hline
    \end{tabular}%
  \label{tab:tab1}%
\end{table*}%

\subsection{Implementation Details}
Our GIC model is built based on the Deit \cite{DEIT} visual transformer-based architecture that is pre-trained at resolution 224$\times$224 and fine-tuned at resolution 384$\times$384 on ImageNet-1k. The Deit backbone encoder consists of L = 12 consecutive transformer blocks with 12 heads, and the patch size is 16. 3 additional learnable transformer blocks with 12 heads are adopted for relation semantic modeling. The dimensions of the visual patch token representations are $D=768$ and projected to a new embedding space with a dimension of $d=512$ (1024 for MSCOCO). The word embedding dimension and the hidden dimension of the grounded language decoder are set to 512 (1024 for MSCOCO). We optimized the proposed model with ADAM \cite{adam} optimizer with a learning rate initialized by 5e-4 (1e-4 for COCO) and annealed by a factor of 0.8 every three epochs. Following existing WSGIC methods \cite{Cyclical, SCAN}, the beam search is disabled for the convenience of grounding evaluation for Flickr30k-Entitie captioning dataset.
\begin{figure}[t!]
\begin{minipage}[b]{1.0\linewidth}
    \centering
    \centerline{\includegraphics[width=7cm]{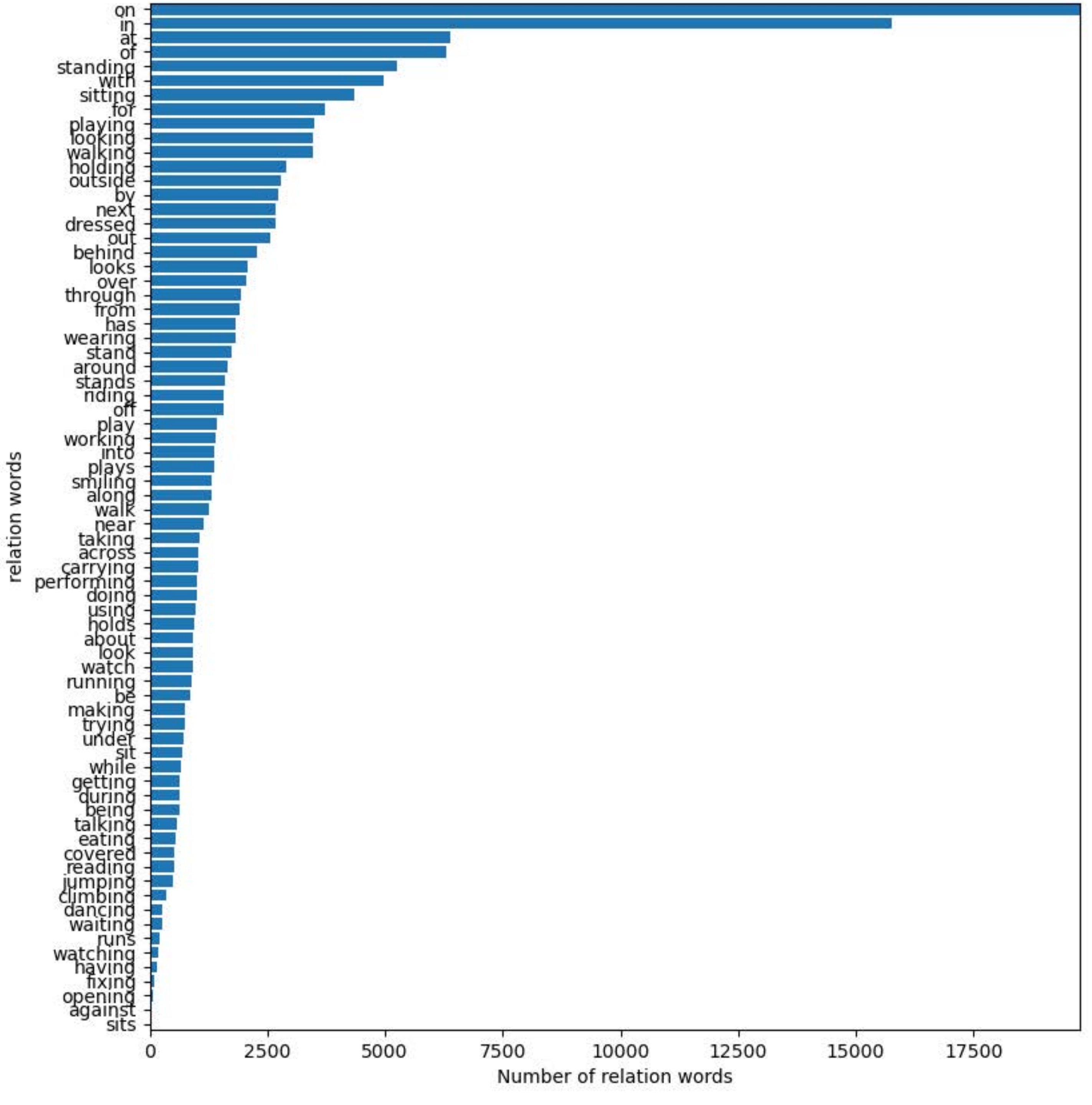}}
    \caption{The statistics of 72 most frequently appearing relation words in the Flickr30K-Entities dataset.}
    \label{fig:flickS}
\end{minipage}
\end{figure}
\begin{figure}[t!]
\begin{minipage}[b]{1.0\linewidth}
    \centering
    \centerline{\includegraphics[width=7cm]{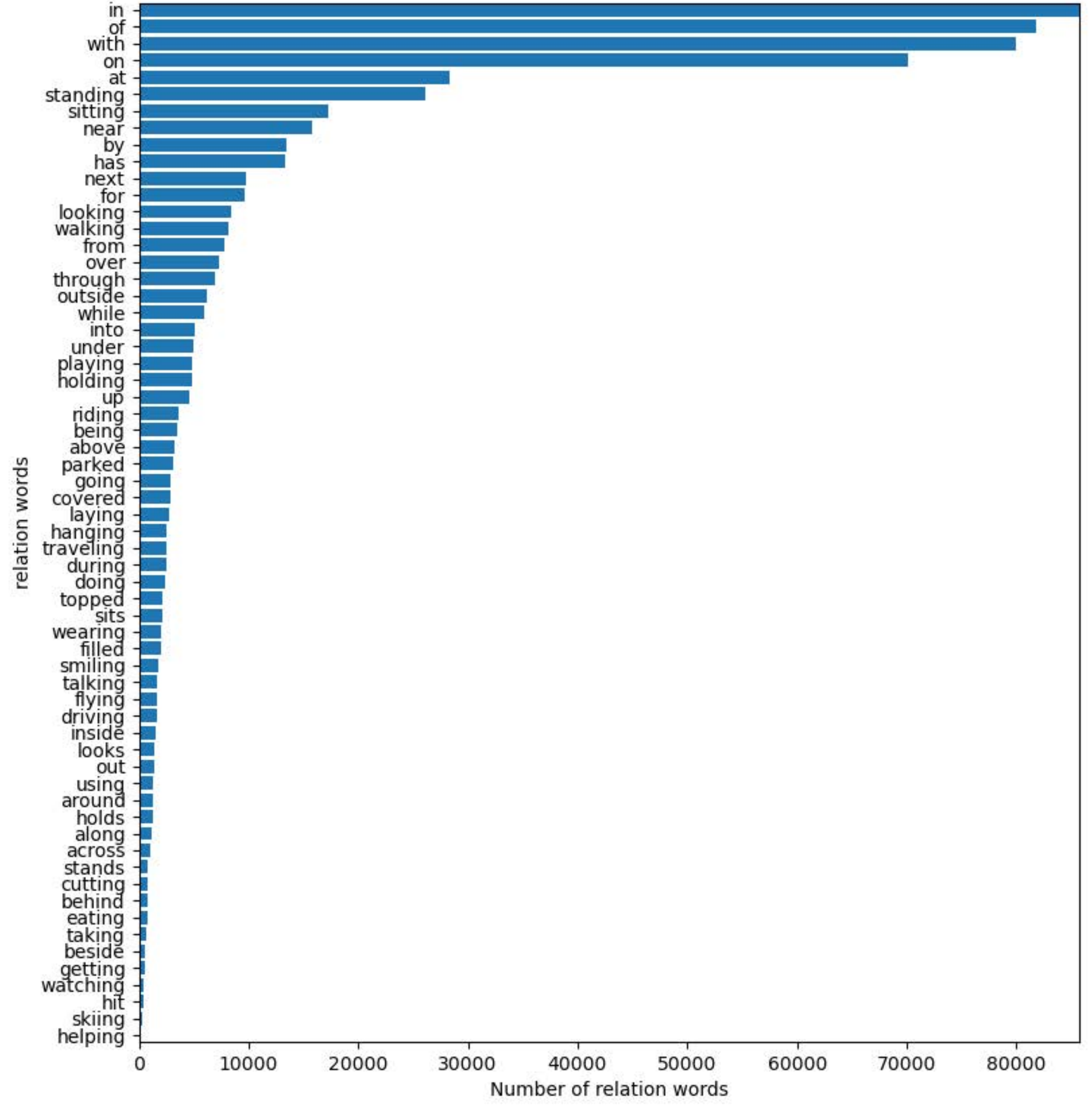}}
    \caption{The statistics of 62 most frequently appearing relation words in the MSCOCO captioning dataset.}
    \label{fig:cocoS}
\end{minipage}
\end{figure}

\subsection{Performance Comparison}
We compared the proposed method with state-of-the-art weakly-supervised grounded image captioning methods on the Flickr-30k test set in Table \ref{tab:tab1}. The comparison includes GVD \cite{GVD}, Cyclical \cite{Cyclical}, Prophet \cite{Prophet} that trained using cross-entropy loss, and weak supervision methods SCAN \cite{SCAN}, CVAE \cite{CVAE}, where $^{\dagger}$ denotes the model is trained with weak supervision and  KL-divergence loss. The SCAN \cite{SCAN} and $\text{CVAE}_{RL}$ \cite{CVAE} methods are fine-tuned with their proposed weak supervision and SCST \cite{SCST} reward using reinforcement learning (RL). Ticks in the table denote the methods are fine-tuned with RL. We achieved significant improvement on both caption and grounding (\text{$\text{F1}_{all}$} and \text{$\text{F1}_{loc}$}) accuracy. Specifically, for methods trained with XE, our proposed methods achieve the performance of 7.88\% for \text{$\text{F1}_{all}$} score and 19.9\% for \text{$\text{F1}_{loc}$} score, which is 1.18 and 0.7 points higher as compared to CVAE \cite{CVAE}. We also achieve better \text{$\text{F1}_{all}$} score as compared to the GVD \cite{GVD} that trained using bounding box supervision (Sup.). Our method that fined-tuned using RL has outperformed the methods \text{$\text{SCAN}^{\dagger}$} and \text{$\text{CVAE}_{RL}^{\dagger}$} that fine-tuned using RL and weak supervision with KL loss. 
Nevertheless, from this table, we observe that optimizing the model with SCST reward could disrupt VLAMs for grounding. 
This disruption is due to the lack of weak supervision \cite{SCAN, CVAE} from region features \cite{RCNN} during the RL training process.
We intend to address this challenge by exploring a solution that eliminates the need for preprocessed region features within the RL framework in future implementations.
It's important to note that the primary focus of this work centers on generating accurate visual-language representation (VLAMs) for grounding in a one-stage framework.
The increments shown in the table have proven the effectiveness of our proposed WSGIC methods.

\begin{table}[t!]
\setlength{\tabcolsep}{9pt}
  \centering
  \caption{Evaluation results for various components used in the model. Performance measured on Flickr30k-Entities test split in (\%).}
    \begin{tabular}{l|cc|cc}
    \hline
    \multicolumn{1}{l|}{Models} & \multicolumn{2}{c|}{Caption eval}     & \multicolumn{2}{c}{Grounding eval} \\
    \hline
           & \multicolumn{1}{c}{B@4} &  \multicolumn{1}{c|}{C}  & \multicolumn{1}{c}{$\text{F1}_{all}$} & \multicolumn{1}{c}{$\text{F1}_{loc}$} \\
    \hline
    $\text{GM}$     & 27.0 & 62.1  & 5.36  & 14.1  \\
    $\text{GM + cls}$     & 27.2 & 62.3   &  5.86  & 15.0 \\
    $\text{GM + tokens}$   & 27.8  &  63.9  &  6.33 & 15.5 \\
    $\text{RGM}$       & 27.5 & 63.8  & 7.34 & 18.7 \\
    $\text{RGM + cls}$      & 28.2 & 63.9  & 7.54 & 18.9 \\
    $\text{RGM + tokens}$      & 27.8 & 64.1  & \textbf{7.88} & \textbf{19.9} \\
    \hline
    \end{tabular}%
  \label{tab:tab2}%
\end{table}%

\begin{table}[t!]
\setlength{\tabcolsep}{6.5pt}
  \centering
  \caption{Results of utilizing different Vision Transformer backbone for captioning and grounding.}
    \begin{tabular}{l|c|cc|cc}
    \hline
    \multicolumn{1}{l|}{Models} & \multicolumn{1}{l|}{Backbone} & \multicolumn{2}{c|}{Caption eval}     & \multicolumn{2}{c}{Grounding eval} \\
    \hline
           &  &  \multicolumn{1}{c}{B@4} &  \multicolumn{1}{c|}{C}  & \multicolumn{1}{c}{$\text{F1}_{all}$} & \multicolumn{1}{c}{$\text{F1}_{loc}$} \\
    \hline
    $\text{RGM}$       & $\text{DeiT-B}$ & 27.5 & 63.8 & \textbf{7.34} & \textbf{18.7} \\
    $\text{RGM}$       & $\text{ViT-B}$ & 27.1  &  63.2  & 5.45 & 14.7 \\
    $\text{RGM}$       & $\text{SwinT-B}$ & 27.7  &  66.0 & 4.15 & 10.3 \\
    \hline
    \end{tabular}%
  \label{tab:tab3}%
\end{table}%


\begin{figure*}[t!]
  \centering
    
    \includegraphics[width=17cm]{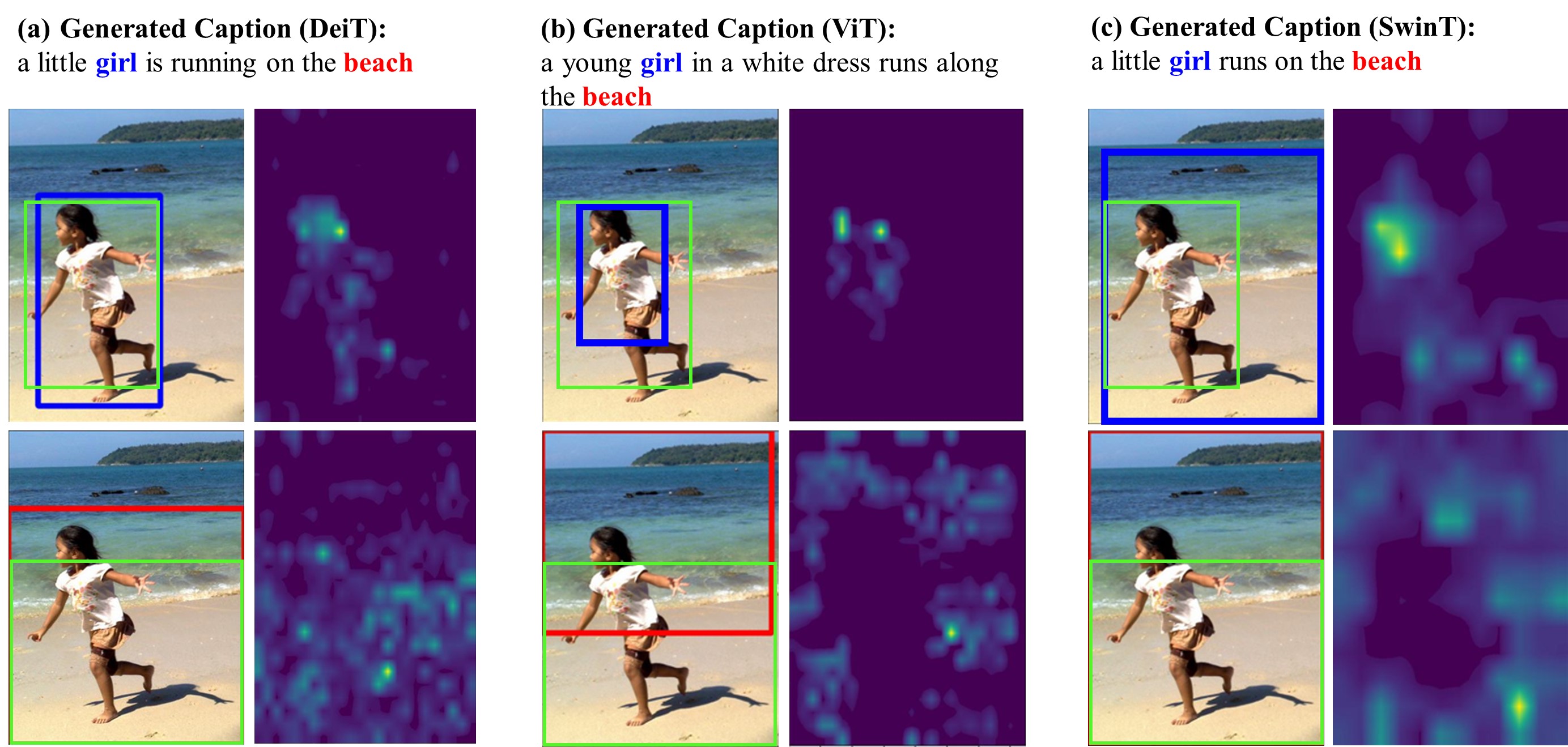}
  
   \caption{Visualization of the generated VLAMs using different backbone networks. The image on the left shows the localized region, whereas the corresponding VLAM is on the right. GT caption: a young girl running on the beach. In the caption, the highlighted object words correspond to the generated colored bounding boxes (blue, red) in the image, with the green bounding box representing the ground truth bounding box.}
   \label{fig:onecol}
\end{figure*}

\subsection{Ablation study}
We conduct ablation studies to verify the effectiveness of various components in the proposed method.
In Table \ref{tab:tab2}, we compare our model in various settings.
The $\text{GM}$ injects visual patch representations $\mathbf{Z}_{\text{patch}}^{L}$ only to the LSTM network to generate word representations $\mathbf{H}$ and use Cross-MHA \cite{Trans} as the Grounding Module (GM) for the computation of VLAMs. $\text{GM+cls}$ injects [CLS] token to the LSTM decoder for word representation generation, and  $\text{GM+tokens}$ includes both [CLS] and [REL] tokens to the LSTM network. The RGM model tests the effectiveness of utilizing fusion features $\mathbf{U}$ to generate VLAMs, and both tokens are injected into the LSTM for word representation generation. For $\text{RGM}_{cls}$, we include [CLS] token in the RGM, which helps predict more accurate groundable words in the caption. In $\text{RGM}_{cls+rel}$, we include both [CLS] and [REL] tokens, which archives the best captioning and grounding performance. Our experiment indicates that the RGM module greatly aids one-stage caption grounding.  Incorporating [CLS] and [REL] tokens can enhance the performance of the grounded captioner, leading to the generation of more precise caption words (resulting in higher B@4 and C scores) and, consequently, improved grounding performance. $\text{RGM}_{cls+rel}$ improves 2.52 and 5.8 points in \text{$\text{F1}_{all}$} and \text{$\text{F1}_{loc}$} score respectively as compared to the baseline $\text{GM}$ model.

Since visual representation plays a critical role in WSGIC, 
we investigated several visual backbones, such as Swin Transformer (SwinT-B) \cite{Swin}, ViT-B \cite{VIT}, and DeiT-B \cite{DEIT} to determine their suitability for the WSGIC model (shown in Table \ref{tab:tab3} and Figure \ref{fig:onecol}). We observed that the generated VLAMs are more accurate for grounding when utilizing the DeiT-B as the backbone network.
The VLAMs for the SwinT tend to focus on the discriminative region of each shift window  \cite{Swin} that affects the generation of VLAMs for grounding. The generated VLAMS with the ViT backbone is slightly less accurate as compared to using the DeiT backbone.

\begin{table}[t!]
\setlength{\tabcolsep}{7.5pt}
  \centering
  \caption{Evaluation results for the interaction between different decoders, backbones, semantic tokens, and grounding module that benefit the grounded captioning model.}
    \begin{tabular}{l|cc|cc}
    \hline
    \multicolumn{1}{l|}{Models} & \multicolumn{2}{c|}{Caption eval}     & \multicolumn{2}{c}{Grounding eval} \\
    \hline
          & \multicolumn{1}{c}{B@4} &  \multicolumn{1}{c|}{C}  & \multicolumn{1}{c}{$\text{F1}_{all}$} & \multicolumn{1}{c}{$\text{F1}_{loc}$} \\
    \hline
    DeiT+LSTM+GM & 27.0 & 62.1  & 5.36  & 14.1  \\
    DeiT+TR+GM & 27.6 & 63.0  & 5.02  & 12.9  \\
    $\text{Detector+LSTM+GM}$     & 28.1 & 61.1   &  6.21  & 15.6 \\
    $\text{RGM}$     & 27.5 & 63.8  & 7.34 & 18.7 \\
    $\text{RGM+tokens}$   & 27.8 & 64.1  & \textbf{7.88} & \textbf{19.9} \\
    \hline
    \end{tabular}%
  \label{tab:tab4}%
\end{table}%

\begin{table}[t!]
\setlength{\tabcolsep}{9pt}
\centering
  \centering
  \caption{The impact of using different numbers of the parallel heads ($N_h$) to compute the VLAMs for grounding. Performance is measured on the Flickr30k-Entities test split.}
    \begin{tabular}{c|ccc|cc}
    \hline
    \multicolumn{1}{c|}{Heads} & \multicolumn{3}{c|}{Caption eval}     & \multicolumn{2}{c}{Grounding eval} \\
    \hline
         & \multicolumn{1}{c}{B@4} & \multicolumn{1}{c}{M} & \multicolumn{1}{c|}{C}  & \multicolumn{1}{c}{{$\text{F1}_{all}$}} & \multicolumn{1}{c}{$\text{F1}_{loc}$} \\
    \hline
    1     & 26.6 &   22.3   & 61.2  & 7.17  & 17.7 \\
    4     & 27.2 &   22.6   & 65.4  & 7.29  & 18.9 \\
    8     & 27.8 &   22.7   & 64.1  & \textbf{7.88}  & \textbf{19.9} \\
    12    & 26.8 &   22.6   & 64.1  & 7.39 & 18.6 \\
    \hline
    \end{tabular}%
    \label{tab:tab5}%
\end{table}%

\begin{table}[t!]
\setlength{\tabcolsep}{9pt}
\centering
  \centering
  \caption{The performance of using different numbers of Transformer layers for relation semantic modeling.}
    \begin{tabular}{c|ccc|cc}
    \hline
    \multicolumn{1}{c|}{Layers} & \multicolumn{3}{c|}{Caption eval}     & \multicolumn{2}{c}{Grounding eval} \\
    \hline
         & \multicolumn{1}{c}{B@4} & \multicolumn{1}{c}{M} & \multicolumn{1}{c|}{C}  & \multicolumn{1}{c}{{$\text{F1}_{all}$}} & \multicolumn{1}{c}{$\text{F1}_{loc}$} \\
    \hline
    1     & 27.9 &   23.0  & 64.8  & 6.78  & 17.5 \\
    2     & 28.0 &   22.8   & 65.4  & 6.86  & 17.3 \\
    3     & 27.8 &   22.7   & 64.1  & \textbf{7.88}  & \textbf{19.9} \\
    4    & 28.0 &   22.7   & 64.2  & 7.83 & 18.3 \\
    5    & 28.2 &   22.6   & 64.4  & 7.44 & 18.8 \\
    6    & 28.3 &   22.6   &  64.9 & 7.30 & 17.8\\
    \hline
    \end{tabular}%
    \label{tab:tab6}%
\end{table}%

\begin{table}[htbp]
\setlength{\tabcolsep}{13.5pt}
  \centering
  \caption{The analysis of FPS and Gflop between existing detector-based methods and Our detector-free approach.}
    \begin{tabular}{l|c|c}
    \hline
    Methods & FPS   & Gflop \\
    \hline
    Detector-based \cite{SCAN, Cyclical} & <5    & >850 \\
    \hline
    Ours  & 32.2  & 64.2 \\
    \hline
    \end{tabular}%
  \label{tab:6}%
\end{table}%

\begin{figure*}[t]
\begin{minipage}[b]{1.0\linewidth}
    \centering
    \centerline{\includegraphics[width=15cm]{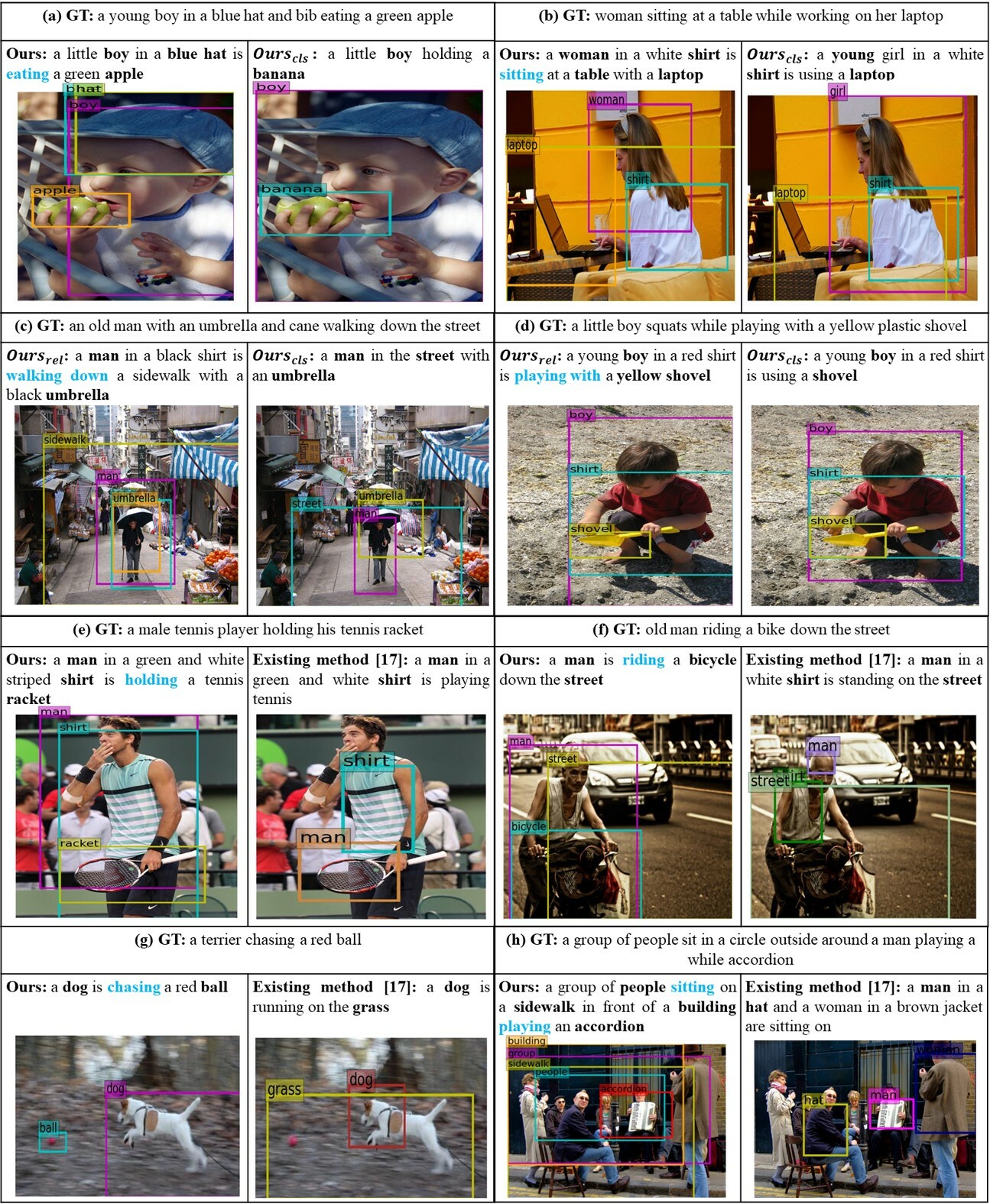}}
    \caption{Examples of generated caption and grounded regions by proposed method and the existing state-of-the-art WSGIC method \cite{SCAN}. Our method generates relation words (blue bold) and groundable object words (black bold) accurately with respect to the ground truth caption (GT).}
    \vspace{-12pt}
    \label{fig:fig7}
\end{minipage}
\end{figure*}

In Table \ref{tab:tab4}, we evaluate results for the interaction between different decoders, backbones, semantic tokens, and grounding modules that benefit the grounded captioning model in localizing more accurate object words. 
The method locates more accurate object words when the \text{$\text{F1}_{all}$} is higher.
The DeiT+TR+GM and DeiT+LSTM+GM indicate we are utilizing the Transformer decoder and the LSTM decoder with DeiT \cite{DEIT} backbone for captioning and measuring the \text{$\text{F1}_{all}$} to evaluate the existence of the correctly grounded object words in the caption, respectively.
GM means that we include the Grounding Module (GM) in the model without using features $\mathbf{U}$. The $\text{Detector+LSTM+GM}$ indicates the region features \cite{RCNN} are adopted for captioning and matched for grounding.
The $\text{RGM}$ denotes we are utilizing the proposed Recurrent Grounding module, and $\text{RGM+tokens}$ injecting both [CLS] and [REL] tokens into the model.
The model with the Transformer decoder tends to have higher captioning evaluation scores (27.6 of B$@$4 and 63.0 of C)  but lower in \text{$\text{F1}_{all}$} (5.02)  than the model using the LSTM decoder. Consequently, given our model's emphasis on computing more accurate VLAMs for object localization, we have chosen LSTM as the baseline decoder for captioning and grounding in this paper. Furthermore, we can see that the captioning performance can be improved by utilizing the RGM module. We observed that the captioning and \text{$\text{F1}_{all}$} score had been significantly improved by utilizing the RGM and semantic tokens, which also implies that the proposed model is prone to predict more accurate object words in the generated caption.

Table \ref{tab:tab5} shows the different number of parallel heads that aggregated to compute the VLAMs for grounding. We achieve the best grounding performance when Heads=8. Table \ref{tab:tab6} shows the performance of additional Transformer layers used to model the relation semantic for better word representations and VLAMs generation. We select the layers of 3 as it perform better in grounding evaluation. The mean Average Precision (mAP) results for multi-label relation class prediction is $46.55\%$. 

We analyze the FPS and Gflop between an existing detector-based method \cite{SCAN, Cyclical, Prophet} and our detector-free based approach in Table \ref{tab:6}. When processing an image, the detector-based captioning methods involve navigating through the backbone network \cite{Swin, resnet}, object detector (e.g., Faster-RCNN with FPS $\sim 5$ \cite{Obj_survey, RCNN}), and captioning model that should have a total FPS $<$ 5. In contrast, our proposed method, which eliminates the need for an object detector, achieves a faster FPS of 32.2. Additionally, the existing methods require a detector, usually demanding an extra 850 Gflops \cite{RCNN, LWR_det}, which is absent in our proposed approach. The calculated total Gflops for our model is 64.2 (including 55 Gflops for Deit \cite{DEIT}). Hence, we can observe that our proposed method is more efficient as compared to existing methods.

\begin{figure}[t!]
  \centering  
  \includegraphics[width=8.5cm]{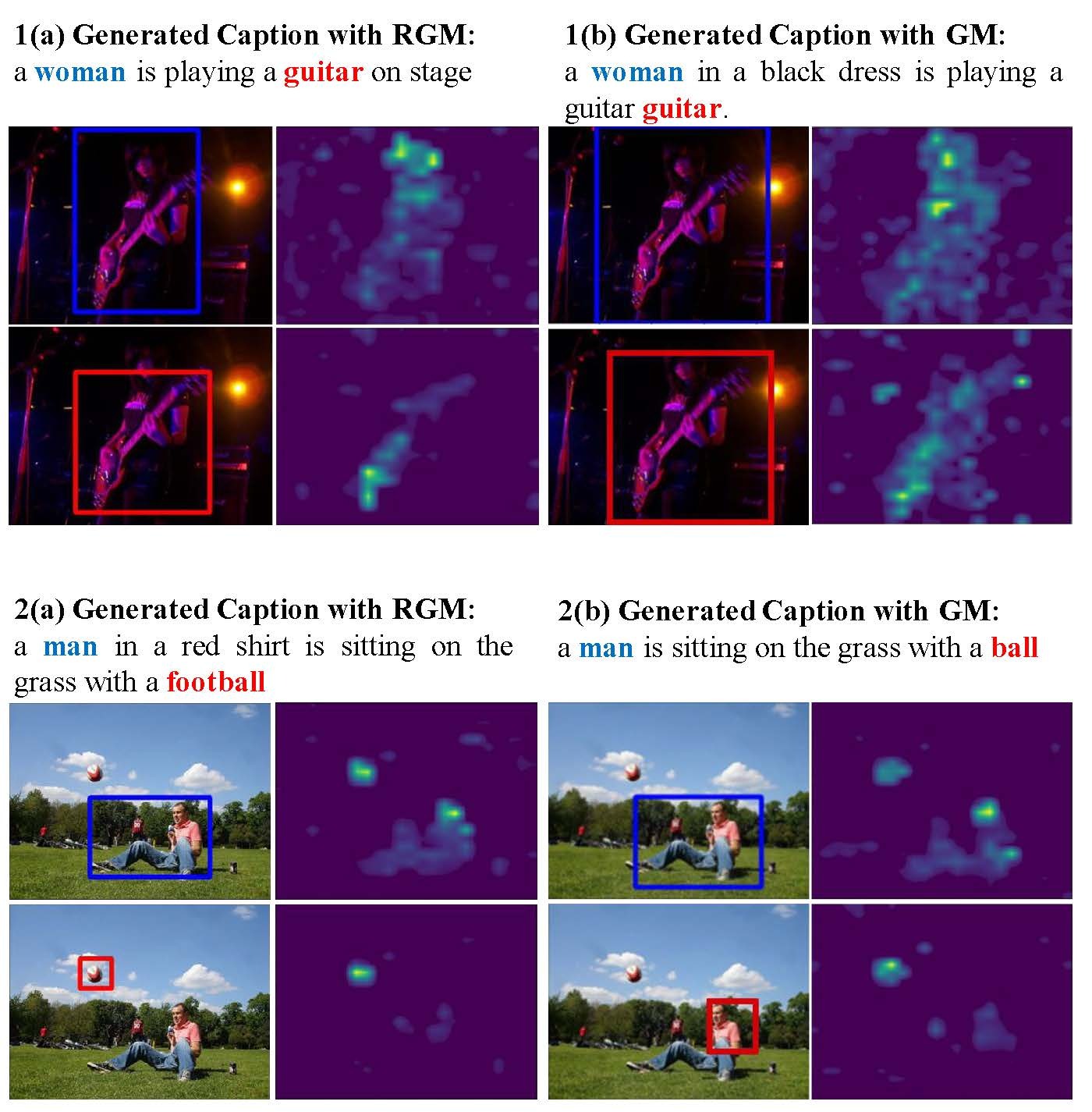}\\

   \caption{Visualization of the generated VLAMs with GM and RGM. To ease
the visualization, we show the VLAMs (on the right) for two object words in the generated caption.}
   \label{fig:fig8}
\end{figure}

\begin{figure*}[t!]
    \begin{center}
        \centering
        \includegraphics[width=17cm]{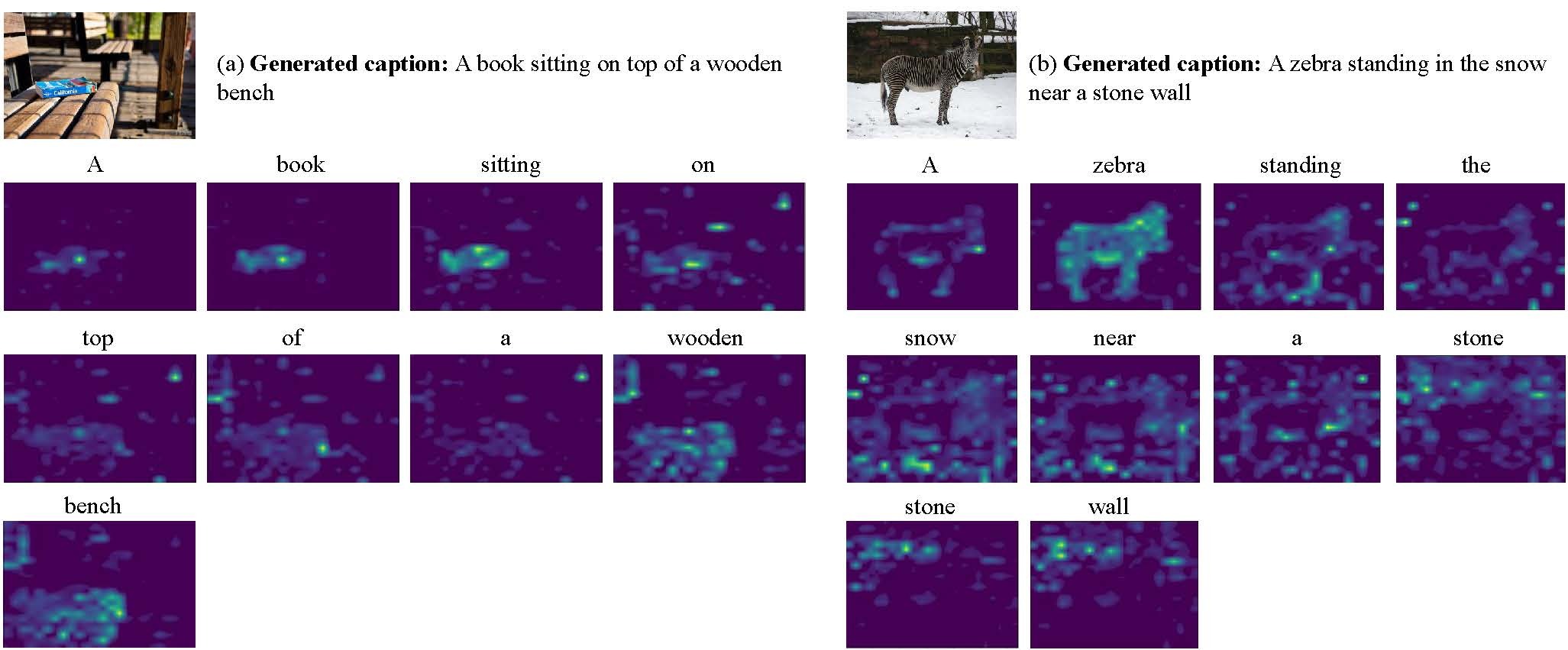}
        \caption{Visualization of the thresholded VLAMs for samples in the MSCOCO captioning test set.}
         \label{fig:fig9}\vspace{-12pt}
    \end{center}
\end{figure*}

\begin{figure}[t!]
  \centering
    \includegraphics[width=8.2cm]{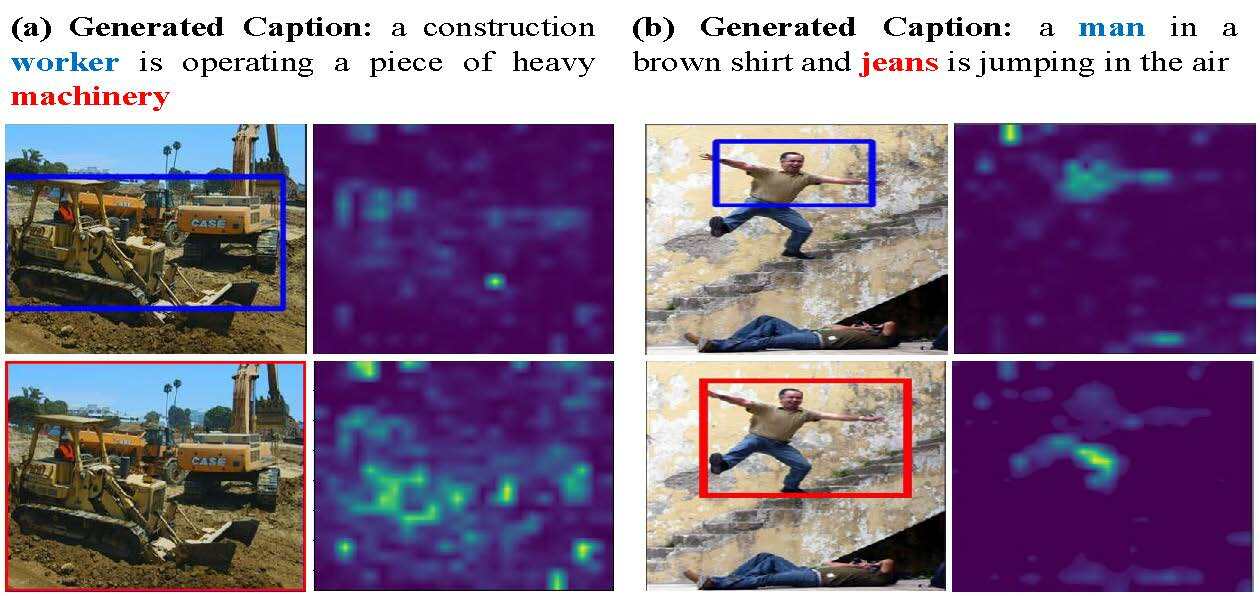}
   \caption{Examples of the failure grounding cases.}
   \label{fig:onecol3}
\end{figure}

\subsection{Qualitative Analysis}
We present some qualitative samples in Figure \ref{fig:fig7}. We compare our proposed method (Ours) using both [REL] and [CLS] tokens and the method ($\text{Ours}_{cls}$) using [CLS] token only in (a) and (b). 
We remark that the method further incorporated with [REL] performs better in generating relation words and can be served as contextual information to aid in the generation of object words in the caption.
 For instance, the model (Ours) correctly predicts the relation words ``eating'' and as context words, leading to predict the correct groundable object words ``apple'' rather than ``banana'' in (a). In (b), it correctly predicts ``sitting at a table'' with respect to the ground truth (GT) caption. 
 In addition, we compare the scenarios of our model that uses [REL] tokens ($\text{Ours}_{red}$) and without using [REL] tokens  ($\text{Ours}_{cls}$) in (c) and (d). 
 We observe that $\text{Ours}_{red}$ tends to predict more accurate relation words, such as "working down" and "playing with," providing context to accurately predict the word "yellow" in (d). 
 However, $\text{Ours}_{red}$ might generate a less accurate object word like "sidewalk" when using only the [REL] token. 
 $\text{Ours}_{cls}$ predicts and locates more accurate object words with a precise bounding box but tends to generate less accurate relation words.  Thus, by leveraging both [REL] and [CLS] tokens, our model can predict more accurate captions and aid in locating precise bounding boxes, as illustrated in (a) and (b).
 Furthermore, we have shown that our proposed models outperform the existing two-stage method \cite{SCAN} in generating correct relation words and object words. For instance, in (e), (g), and (h), our model predicts the correct relation words ``holding'', ``chasing'' and ``playing'' and the more desired object words of ``racket'', ``ball'', and ``accordion'' in the caption with respect to the GT caption. Furthermore, our proposed method successfully grounds the entire regions of the generated groundable object words rather than just part of the body in (e), (f), and (g) (e.g., locate the entire body of a man). 
This qualitative study further proves that our proposed method accurately predicts both relation and groundable object words within the caption, effectively localizing the entire object region within the image. 

  



\begin{table}[t!]
\setlength{\tabcolsep}{9pt}
  \centering
  \caption{Grounding performance for MS-COCO for 80 object classes.}
    \begin{tabular}{l|cc|cc}
    \hline
    \multicolumn{1}{l|}{Models} & \multicolumn{2}{c|}{Caption eval}     & \multicolumn{2}{c}{Grounding eval} \\
    \hline
          & B@4        & C     & \multicolumn{1}{c}{{$\text{F1}_{all}$}} & \multicolumn{1}{c}{$\text{F1}_{loc}$}  \\
    \hline
    $\text{LSTM}$   & 31.3   & 101.4  &  -  & 6.67 \\
    $\text{Detector+GM}$   &  36.8   &  116.2  &  6.26  & 8.13 \\
    $\text{Ours}_{\text{cls}}$  & 36.0   &  116.7  &   7.82    & 9.83  \\
    $\text{Ours}$  &  36.5     &  117.7     &   \textbf{8.78}    &  \textbf{11.5} \\
    \hline
    \end{tabular}%
  \label{tab:tab7}%
\end{table}%

We analyzed the VLAMs generated by GM and proposed RGM in Figure \ref{fig:fig8}. We observed that the generated VLAMs using GM tend to contain more noise as compared to using RGM and result in localizing the regions with noise. For instance, comparing 1(a) and 1(b), the VLAMs generated with RGM in (a) attend to the object itself and yield more precise bounding boxes for object grounding as compared to (b). Furthermore, the VLAM in 2(b) tends to localize (ground the largest contour \cite{thresholding}) the noisy region in the man's body rather than the ball, whereas 2(a) is able to eliminate the noise using RGM and locate the spatial region of football correctly. The incorporation of RGM leads to an enhancement in the grounding performance of the one-stage grounded captioner.

We also present some representative failure cases of our one-stage WSGIC model (shown in Figure \ref{fig:onecol3}). It can be challenging to locate specific small objects in highly complex input images with a weakly-supervised training strategy, as illustrated by the result (a), where there is difficulty grounding the word ``worker'' within the machinery. 
Furthermore, the model may tend to localize more significant regions for an object in the image. 
For example, case (b) localizes the entire human body instead of explicitly locating the ``jeans'', despite the Visual-Linguistic Attention Maps (VLAMs) placing more attention on the jeans. 
These missed localizations result in a lower Intersection over Union (IoU) score, impacting grounding accuracy.

\subsection{Performances on MSCOCO dataset}
Table \ref{tab:tab7} evaluates the grounding performance for the MSCOCO captioning dataset. We evaluate the results under cross-entropy loss training. 
The beam search is disabled for grounding evaluation. 
We evaluate the result using DeiT-B \cite{DEIT}.
To the best of our knowledge, no existing grounding evaluation is available for the MSCOCO captioning dataset. Hence, we evaluate grounding performance with $\text{Detector+GM}$ \cite{RCNN} baseline with our top-down method. 
Our proposed method ($\text{Ours}$) achieves 2.52 and 3.37 points higher in grounding performance. We also compared the proposed method ($\text{Ours}_{cls}$) with using the [CLS] token, and Ours achieves both higher scores in $\text{F1}_{all}$ and $\text{F1}_{loc}$. This further proves the effectiveness of involving the [REL] token in the proposed method that improves the captioning score and predicts more accurate object words in the caption by getting a higher $\text{F1}_{loc}$ score.

\begin{table}[t!]
  \centering
  \setlength{\tabcolsep}{5.5pt}
  \caption{Performance comparison on the MS-COCO Karpathy test split with the weakly-supervised grounded image captioning \cite{SCAN} and recent one-stage $\text{CLIP}$-based \cite{clip} captioning methods. \textcolor{red}{Red} number indicates the best result of using XE loss, \textcolor{blue}{blue} number denotes the best in RL optimization.}
    \begin{tabular}{l|ccccc}
    \hline
    Models  & \multicolumn{1}{c}{B1} & \multicolumn{1}{c}{B4} & \multicolumn{1}{c}{M} & \multicolumn{1}{c}{C} & \multicolumn{1}{c}{S} \\
    \hline
    BUTD \cite{BUTD} & 77.2 & 36.2 & 27.0 & 113.5 & 20.3 \\
    {$+\text{RL}$}\cite{SCST} & 79.8 & 36.3 & 27.7 & 120.1 & 21.4 \\
    \hline
    SCAN\cite{SCAN} & 76.6 & 36.5 & 27.9 & 114.9 & 20.8 \\
    $+\text{RL}$  & 80.2 & 38.0 & 28.5 & 126.1 & 22.0 \\
    \hline
    $\text{PrefixCap}_{\text{ClipViT-B}}$ \cite{PrefixClip}  & - & 33.5 & 27.5 & 113.1 & 21.1 \\
    \hline
    $\text{SmallCap}_{\text{ClipViT-B}}$ \cite{smallcap} & - & 37.0 & 27.9 & 119.1 & 21.3 \\
    \hline
    $\text{CLIPCap}_{\text{ClipViT-B}}$ \cite{clipcap} & - & 37.2 & 28.4 & 119.9 & 21.3 \\
    {$+\text{RL}$} & - & 38.7 & 29.2 & 132.7 & 23.0 \\
    \hline
    $\text{Ours}_{\text{DeiT-B}}$ & 77.2 & 36.5 & 28.2 & 117.7 & 21.4 \\
    $+\text{RL}$ & 79.7 & 38.5  & 28.8 & 126.2 & 22.4 \\
    \hline
    $\text{Ours}_{\text{SwinT-B}}$ & 77.9 & 37.0 & \color{red}\textbf{28.7} & 121.3 & \color{red}\textbf{22.0} \\
    $+\text{RL}$ & 80.4 & 39.6 & 29.3 & 129.7 & 22.8 \\
    \hline
    $\text{Ours}_{\text{ClipViT-B}}$ & \color{red}\textbf{78.2} & \color{red}\textbf{37.4} & \color{red}\textbf{28.7} & \color{red}\textbf{121.7} & 21.9 \\
    $+\text{RL}$ & \color{blue}\textbf{80.9} & \color{blue}\textbf{39.8} & \color{blue}\textbf{29.5} & \color{blue}\textbf{133.2} & \color{blue}\textbf{23.2} \\
    \hline
    \end{tabular}%
  \label{tab:eight}%
\end{table}%

In Table \ref{tab:eight}, we evaluate our captioning performance and compare the result with recent WSGIC \cite{SCAN} and one-stage $\text{CLIP}$ \cite{clip} based image captioning methods \cite{PrefixClip, clipcap, smallcap}. These models are trained with cross-entropy loss and further optimized with RL \cite{SCST}. We compare the results with CLIPCap \cite{clipcap}, PrefixCap \cite{PrefixClip}, and SmallCap \cite{smallcap} using the same $\text{CLIP}_{\text{ViT-B}}$ (Base) backbone for captioning. Our proposed method outperforms the pioneer WSGIC methods \cite{SCAN} in most metrics and achieves competitive performance as compared to one-stage captioning methods \cite{PrefixClip, clipcap, smallcap}. This highlights the effectiveness of our proposed approach. In this paper, we mainly investigate the effectiveness of utilizing a grounding module and [REL] token in generating a more interpretable one-stage grounded image caption, the captioning performance can be further enhanced with more powerful backbones as proven in existing work $\text{CLIPcap}_{\text{ViT-L}}$ \cite{clipcap}.

\subsection{Qualitative results for the MSCOCO captioning dataset} 
In this section, we showcase the generated captions and thresholded Visual Language Attention Maps (VLAMs) from the MSCOCO captioning test split samples in Fig. \ref{fig:fig9}. It is evident that the Visual-Linguistic Attention Maps (VLAMs) generated can effectively align visual and language features, resulting in accurate attention allocation to objects within the image.
For instance, the VLAM successfully directs attention to object words such as ``book" and ``bench" in the generated caption (a), ``A book sitting on top of a wooden bench.'' Likewise, it provides a clear VLAM for ``zebra'' and ``wall'' for the generated caption of ``a zebra standing in the snow near a stone wall.'' 

\section{Conclusion}
In this work, we propose a one-stage weakly-supervised ground image captioning model that generates captions and localizes the groundable words in a top-down manner. We introduce a new token to capture the relation semantic information that serves as context information to eventually benefit the captioning and grounding performance. Furthermore, we propose to compute accurate visual language attention maps (VLAMs) recurrently, which allows it to give a higher quality grounding for the groundable object words. Experiments of two datasets show that the proposed method achieves state-of-the-art performance on captioning and grounding.

\balance
\FloatBarrier

\bibliographystyle{IEEEtran}
\bibliography{reference}

\end{document}